\documentclass[runningheads,a4paper]{llncs}

\usepackage{makeidx}  % allows for indexgeneration
\usepackage{tablefootnote}

\usepackage{amssymb}
\usepackage{multirow}
\usepackage{amsfonts}
\usepackage{graphicx}
\usepackage{pbox}

\begin{document}

\pagestyle{headings}

\title{Discovering and Exploiting Entailment Relationships in Multi-Label Learning}
\titlerunning{Discovering \& Exploiting Entailment Relationships in Multi-Label Learning}

\author{Christina Papagiannopoulou\inst{1} \and Grigorios Tsoumakas\inst{1} \and \\ Ioannis Tsamardinos\inst{2,3}}

\authorrunning{C. Papagiannopoulou \and G. Tsoumakas \and I. Tsamardinos}

\institute{Aristotle University of Thessaloniki, Thessaloniki 54124, Greece\\
\email{cppapagi,greg@csd.auth.gr}%\\ WWW home page: \texttt{http://users.auth.gr/\homedir greg}
\and
Computer Science Department, University of Crete, Greece\\
\and
Institute of Computer Science, Foundation for Research and Technology - Hellas, \\
N. Plastira 100 Vassilika Vouton, GR-700 13 Heraklion, Crete, Greece \\
\email{tsamard@ics.forth.gr}}
\maketitle

\begin{abstract}
This work presents a sound probabilistic method for enforcing adherence of the marginal probabilities of a multi-label model to automatically discovered deterministic relationships among labels. In particular we focus on discovering two kinds of relationships among the labels. The first one concerns pairwise positive entailment: pairs of labels, where the presence of one implies the presence of the other in all instances of a dataset. The second concerns exclusion: sets of labels that do not coexist in the same instances of the dataset. These relationships are represented with a Bayesian network. Marginal probabilities are entered as soft evidence in the network and adjusted through probabilistic inference. Our approach offers robust improvements in mean average precision compared to the standard binary relavance approach across all 12 datasets involved in our experiments. The discovery process helps interesting implicit knowledge to emerge, which could be useful in itself. 
\end{abstract}

\section{Introduction}

Learning from multi-label data has received a lot of attention from the machine learning and data mining communities in recent years. This is partly due to the multitude of practical applications it arises in, and partly due to the interesting research challenges it presents, such as exploiting label dependencies, learning from rare labels and scaling up to large number of labels \cite{tsoumakas+etal:2012}.

%could delete reference to zhang and to special issue.

In several multi-label learning problems, the labels are organized as a tree or a directed acyclic graph, and there exist approaches that exploit such structure \cite{vens+etal:2008,barutcuoglu+etal:2006}. However, in most multi-label learning problems, flat labels are only provided without any accompanying structure. Yet, it is often the case that implicit deterministic relationships exist among the labels. For example, in the ImageCLEF 2011 photo annotation task, which motivated the present study, the learning problem involved 99 labels\footnote{\url{http://www.imageclef.org/system/files/concepts\_2011.txt}} without any accompanying semantic meta-data, among which certain deterministic relationships did exist. For example, among the 99 labels were several groups of mutually exclusive labels, such as the four seasons {\em autumn,  winter, spring, summer} and the person-related labels {\em single person, small group, big group, no persons}. It also included several positive entailment (consequence) relationships, such as $river \rightarrow  water$ and $car \rightarrow vehicle$. Hierarchies accompanying multi-label data model positive entailment via their is-a edges, but do not model exclusion relationships. 

These observations motivated us to consider the automated learning of such deterministic relationships as potentially interesting and useful knowledge, and the exploitation of this knowledge for improving the accuracy of multi-label learning algorithms. While learning and/or exploiting {\em deterministic} relationships from multi-label data is not new \cite{park+fuernkranz:2008}, little progress has been achieved in this direction since then. Past approaches exhibit weaknesses such as being unsuccesful in practice \cite{park+fuernkranz:2008}, lacking formal theoretical grounding \cite{mbanya+etal:2010,mbanya+etal:2011} and being limited to existing is-a relationships \cite{barutcuoglu+etal:2006}. 

Given an unlabeled instance $x$, multi-label models can output a bipartition of the set of labels into relevant and irrelevant to $x$, a ranking of all labels according to relevance with $x$, marginal probabilities of relevance to $x$ for each label or even a joined probability distribution for all labels. The latter is less popular due to the exponential complexity it involves \cite{dembczynski:2010a}. Among the rest, marginal probabilities are information richer, as they can be cast into rankings after tie breaking and into bipartitions after thresholding. They are also important if optimal decision making is involved in the application at hand, which is often the case.  

This work presents a sound probabilistic method for enforcing adherence of the marginal probabilities of a multi-label model to automatically discovered deterministic relationships among labels. We focus on two kinds of relationships. The first one concerns pairwise {\em positive entailment}: pairs of labels, where presence of one label implies presence of the other in all instances of a dataset. The second one concerns {\em exclusion}: sets of labels that do not coexist at the same instances of a dataset. These relationships are represented with a Bayesian network. Marginal probabilities are entered as soft evidence in the network and adjusted through probabilistic inference. Our approach offers robust improvement in mean average precision compared to the standard binary relavance approach across all 12 datasets involved in our experiments. The discovery process helps interesting implicit knowledge to emerge, which could be useful in itself. 

The rest of this paper is organized as follows. Section 2 introduces our approach. In particular, Section 2.1 discusses the discovery and Section 2.2 the exploitation of entailment relationships. Section 3 presents related work and contrasts it with our approach. Section 4 presents the empirical work, with Section 4.1 discussing datasets and experimental setup, Section 4.2 presenting samples of the knowledge discovered by our approach and Section 4.3 discussing comparative prediction results against binary relevance. Finally, Section 5 summarizes the conclusions of this work and suggests future work directions. 

\section{Our Approach}

\subsection{Discovering Entailment Relationships}

Let $A$ and $B$ be two labels with domain $\{false, true\}$. For simplicity, we will be using the common shortcut notation $a$, $\neg a$, $b$ and $\neg b$ instead of $A=true$, $A=false$, $B=true$ and $B=false$ respectively. The following four entailment relationships can arise between the two labels: 

\begin{enumerate}
\item $a \rightarrow b$ and equivalent contrapositive $\neg b \rightarrow \neg a$ (positive entailment)
\item $b \rightarrow a$ and equivalent contrapositive $\neg a \rightarrow \neg b$ (positive entailment)
\item $a \rightarrow \neg b$ and equivalent contrapositive $b \rightarrow \neg a$ (exclusion)
\item $\neg a \rightarrow b$ and equivalent contrapositive $\neg b \rightarrow a$ (coexhaustion)
\end{enumerate}

Figure \ref{fig:contingency} presents a contingency table for labels $A$ and $B$, based on a multi-label dataset with $S+T+U+V$ training examples. Positive entailment corresponds to $T=0$ or $U=0$, exclusion to $S=0$ and coexhaustion to $V=0$. Furthermore, $S = V = 0$ corresponds to mutually exclusive and completely exhaustive labels, while $T = U = 0$ corresponds to equivalent labels. 

\begin{figure}
\centering
\setlength{\tabcolsep}{10pt}
\renewcommand{\arraystretch}{1.3}
\begin{tabular}{c|c|c}
          & $a$ & $\neg a$ \\
\hline
 $b$      & $S$ & $T$ \\
\hline
 $\neg b$ & $U$ & $V$ \\ 
\hline
\end{tabular}
\caption{Contingency table for labels $A$ and $B$}
\label{fig:contingency}
\end{figure}

In this work, we focus on discovering {\em pairwise} positive entailment relationships as well as exclusion relationships among {\em two or more} labels. For a multi-label dataset with $q$ labels, it is easy to extract all four types of pairwise entailment relationships from the corresponding contingency tables in $O(q^2)$ time complexity. For discovering exclusion relationships among more than two labels, we follow the paradigm of the Apriori algorithm \cite{agrawal+rankrishnan:1994} in order to find all maximal sets of mutually exclusive labels, such that each of them is not a subset of another. Starting from the pairwise exclusion relationships, we find triplets of mutual exclusive labels, then quads and so on. 

%discuss pruning of relationships based on threshold

As a toy example, consider the label values of a multi-label dataset with 6 labels that are given in Table \ref{tbl:discovery-example}, where to improve legibility we have used a value of 1 to represent {\em true} and a value of 0 to represent {\em false}. Our approach would in this case extract the positive entailment relationships $a \rightarrow b$, $a \rightarrow c$, $b \rightarrow c$ and $d \rightarrow c$, and an exclusion relationship for the set of labels $\{A, E, F\}$.

\begin{table*}
\setlength{\tabcolsep}{10pt}
\centering
\caption{A toy multi-label dataset with 10 samples and 6 labels.}
\begin{tabular}{llllll}
\hline\noalign{\smallskip}
A & B & C & D & E & F \\
\hline\noalign{\smallskip}
1 & 1 & 1 & 0 & 0 & 0 \\
1 & 1 & 1 & 1 & 0 & 0 \\
0 & 0 & 0 & 0 & 1 & 0 \\
0 & 1 & 1 & 0 & 1 & 0 \\
1 & 1 & 1 & 0 & 0 & 0 \\
0 & 1 & 1 & 1 & 0 & 1 \\
0 & 0 & 1 & 1 & 1 & 0 \\
0 & 0 & 0 & 0 & 0 & 1 \\
0 & 0 & 1 & 0 & 0 & 0 \\
0 & 0 & 0 & 0 & 0 & 1 \\
\noalign{\smallskip}\hline
\end{tabular}
\label{tbl:discovery-example}
\end{table*}

%An important issue concerns the minimum number of examples that should support each relationship, in order for it to be considered valid. 

\subsection{Exploiting Entailment Relationships}

Motivated by the goal of theoretically sound correction of the marginal probabilities $P_x(\lambda)$, obtained by a multi-label model for each label $\lambda$ given instance $x$, according to the background knowledge expressed by the discovered relationships, we propose using a Bayesian network for the representation of these relationships. This network initially contains as many nodes as the labels, with each node representing the conditional probability $P_x(\lambda)$ of corresponding label $\lambda$, given instance $x$, with uniform prior. 

We represent the entailment relationship $a \rightarrow b$, among labels $A$ and $B$, by adding a link from node $A$ to node $B$. We set the conditional probability table (CPT) associated with node $B$ to contain probabilities $P_x(b|a)=1$ and $P_x(b|\neg a)=0$. This is easily generalized in the case of multiple relationships $a_1 \rightarrow b, \ldots, a_k \rightarrow b$, to a CPT with $P_x(b|A_1, \ldots, A_k)=1$ if $A_1 \vee \ldots \vee A_k = true$ and $P_x(b|A_1, \ldots, A_k)=0$ otherwise (i.e. when $A_1 \vee \ldots \vee A_k = false$ or equivalently when $\neg A_1 \wedge \ldots \wedge \neg A_k = true$). Such a CPT renders node $B$ deterministic. Our representation assumes that all causes of $B$ have been considered, which is seldom true for typical multi-label datasets. To deal with this discrepancy, we add an additional parent of $B$ as {\em leak} node, corresponding to a new virtual label whose value is set to: (i) {\em false} in those training examples where $B=false$, (ii) {\em true} in those training examples where $B=true$ and all other parents of $B$ are {\em false}, and (iii) {\em false} for the rest of the training examples. Note that for the last category of examples where $B=true$ and at least one other parent is {\em true}, the value of the leak node could also be set to {\em true} instead of false, but the choice of {\em false} should lead to semantically simpler virtual labels that are easier to learn. Redundant relationships due to the transitivity property of positive entailment are not represented in the network.

%must check how seldom exhaustive relationships are in practice

%discuss alternative handling

We represent the mutual exclusion relationship among labels $A_1, \ldots, A_k$ by adding a new boolean deterministic node $B$ as common child of all of these labels. We set the CPT of this node to contain probabilities $P_x(b|A_1, \ldots, A_k)=1$ if one and only one of the parents is true and $P_x(b|A_1, \ldots, A_k)=0$ otherwise. We consider that this node is true as observed evidence ($B=true$). Our representation assumes that labels $A_1, \ldots, A_k$ cover all training examples, which usually is not the case for typical multi-label datasets. We deal with this discrepancy similarly to the case of positive entailment. In specific, we add an additional parent of $B$ as {\em leak} node, corresponding to a new virtual label whose value is set to: (i) {\em true} in those training examples where all other parents of $B$ are {\em false}, and (ii) {\em false} in all other training examples. 

\begin{figure}
\centering
\includegraphics[scale=0.4]{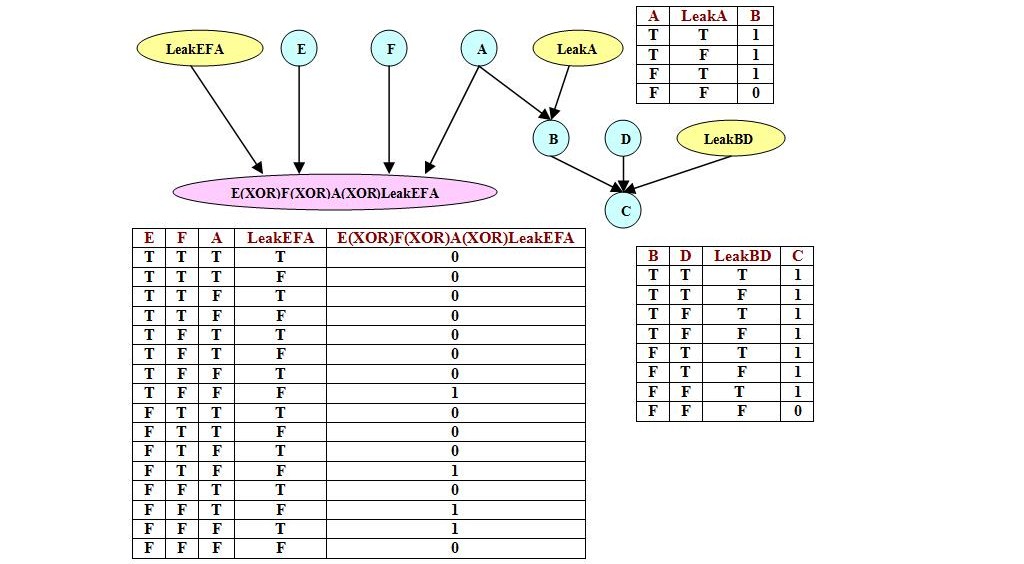}
\caption{A network that represents labels $A, B, C, D, E, F$, entailment relationships $a \rightarrow b$, $a \rightarrow c$, $b \rightarrow c$, $d \rightarrow c$ and a mutual exclusion relationship among labels $A, E$ and $F$.}
\label{fig:network}
\end{figure}

Continuing the toy example of the previous section, Figure \ref{fig:network} shows the network that our approach would construct to represent the discovered knowledge.

%labels $A, B, C, D, E, F$, positive entailments $a \rightarrow b$, $a \rightarrow c$, $b \rightarrow c$, $d \rightarrow c$ and a mutual exclusion among labels $A, E$ and $F$.

Having constructed the network, we then use any multi-label algorithm that can provide marginal probability estimates to fit the extended (with virtual labels corresponding to leak nodes) training set. For an unlabelled instance $x$, we first query the multi-label model in order to obtain probability estimates $P_x(\lambda)$ for each of the labels $\lambda$, including virtual labels. These are then entered into the network as {\em soft} (also called {\em virtual}) evidence \cite{korb+nicholson:2003}. A probabilistic inference algorithm is then used to update the probability estimates for each of the labels, leading to probabilities that are consistent with the discovered relationships. %The network is then reset to its prior state in order to process a subsequent unlabelled instance. 

Table \ref{tbl:exampleUpdating} exemplifies the probability correction process. Each column corresponds to a particular label/node. The first row contains arbitrary probability estimates for an instance. These probabilities  violate the relationships we have discovered. The probability of label $C$ should be larger than that of $B$, $D$ and $LeakBD$, but only the last constraint holds. In addition the probability of $B$ should be larger than that of $A$ and $LeakA$, none of which holds. Finally, the probabilities of $A$, $E$, $F$ and $LeakEFA$ should add to 1, which does not hold. The third row gives the adjusted probabilities according to our approach, which are now consistent with the discovered relationships.

\begin{table}
\setlength{\tabcolsep}{4.7pt}
\centering
\caption{Marginal probabilities obtained by the multi-label model for each of the labels, including the virtual ones corresponding to leak nodes (before) and updated probabilities after probabilistic inference (after) for  the network in Figure~\ref{fig:network}.}
\begin{tabular}{cccccccccc}
\hline\noalign{\smallskip}
&  $A$ & $LeakA$& $B$ & $D$ & $LeakBD$ & $C$  & $F$ & $E$ & $LeakEFA$ \\
\hline\noalign{\smallskip}
Before & 0.4  & 0.35  & 0.25  & 0.6     & 0.01 & 0.2   & 0.3   & 0.85 & 0.3 \\
After  & 0.022 & 0.082 & 0.096 & 0.031  & 0.05 & 0.345 & 0.064 & 0.85 & 0.064 \\ 
\hline
\end{tabular}\\
\label{tbl:exampleUpdating}
\end{table}
                                                              
\section{Related Work}

The idea of discovering and exploiting label relationships from multi-label data was first discussed in \cite{park+fuernkranz:2008}, where relationships were reffered to as {\em constraints}. An interesting general point of \cite{park+fuernkranz:2008} was that label constraints can be exploited either at the learning phase or at a post-processing phase. In addition, it presented four basic types of constraints, which correspond to the four entailment relationships, and noted that more complex types of constraints can be represented by combining these basic constraints with logical connectors. For  discovering constraints, it proposed association rule mining, followed by  removal of redundant association rules that were more general than others. For exploiting constraints, it proposed two post-processing approaches for the label ranking task in multi-label learning. These approaches correct a predicted ranking when it violates the constraints by searching for the nearest ranking that is consistent with the constraints. They only differ in the function used to evaluate the distance between the invalid and a valid ranking. As they focus on label ranking, these approaches cannot be used for correcting the marginal probability estimates of the labels. Results on synthetic data with known constraints showed that constraint exploitation can be helpful, but results on real-world data and automatically discovered constraints did not lead to predictive performance improvements. 
%The authors acknowledged professor Tom Mitchell for suggesting this particular learning setting!

An approach for exploiting label relationships in order to update the marginal probabilites was presented in \cite{mbanya+etal:2010,mbanya+etal:2011}. For mutually exclusive labels that cover all training examples, the idea was to keep the highest probability and set the rest of the marginal probabilities to zero. For mutually exclusive labels that don't cover all training examples, this rule was used if the highest probability was larger than a threshold. For an entailment relation $a \rightarrow b$, the idea was to set the marginal probability of $b$ to that of $a$ when the marginal probability of $a$ is larger than that of $b$ and larger than 0.5. This post-processing phase approach, called {\em concept reasoning}, leads to marginal probabilities that are consistent with the label relationships, but lacks a sound probabilistic inference framework. 

The exploitation of parent-child relationships of a {\em given} hierarchical taxonomy of labels via a Bayesian network structure was proposed in \cite{barutcuoglu+etal:2006}. In particular, parent labels were conditioned on their child labels, and the output of the binary classifier for each label was conditioned on the corresponding label. Given the classifier outputs as evidence, a probabilistic inference algorithm is then applied to obtain hierarchically consistent marginal probabilities for each label. Our approach shares the same principle of using a Bayesian network structure for enforcing consistency of marginal probabilities, but: (i) constructs the structure automatically according to relationships discovered from the data, (ii) represents mutual exclusion relationships in addition to positive entailment, and (iii) builds additional binary models for virtual labels corresponding to leak nodes in the deterministic relations represented by the network. 

A Bayesian network structure to encode the relationships among labels as well as between the input attributes and the labels was presented in \cite{zhang+zhang:2010}. The proposed algorithm, called LEAD, starts by building binary relevance models and continues by learning a Bayesian network on the residuals of these models. Then another set of binary models is learned, one for each label, but this time incorporating the parents of each label according to the constructed Bayesian network as additional features. For prediction, these models are queried top-down according to the constructed Bayesian network. LEAD implicitly discovers probabilistic relationships among labels by learning the Bayesian network structure directly from data, while our approach explicitly discovers deterministic positive entailment and mutual exclusion relationships among labels from the data, which are then used to define the network structure. 

A method for uncovering deterministic causal structures is introduced in \cite{baumgartner:2009}. Similarly to our work, it aims at constructing a Bayesian network out of automatically discovered deterministic relationships. Important differences are that it does not consider latent variables, as in our representation of exclusions and our treatment of unaccounted causes of a label via leak nodes. It therefore requires relationships to be supported from the full dataset, which limits its practical usefulness, as rarely such relationships appear in real-world data.

%http://dx.doi.org/10.1016/j.robot.2013.10.001

%references found by Christina on entailment

\section{Experiments}

\subsection{Setup}

We use the binary relevance (BR) problem transformatiom method for learning multi-label models, which learns one binary model per label. As our approach employs probabilistic inference to {\em correct} the marginal probability of each label, we consider important to start with good probability estimates. We therefore use Random Forest \cite{breiman:2001} as the learning algorithm for training each binary model, since it is known to provide relatively accurate probability estimates without calibration \cite{Niculescu-Mizil+Caruana:2005}. We use the implementations of BR and Random Forest from Mulan \cite{tsoumakas+etal:2011b} and Weka \cite{hall+etal:2009} respectively. Our approach is also implemented in Mulan, utilizing the jSMILE library\footnote{\url{http://genie.sis.pitt.edu/}}.% for Bayesian network representation and probabilistic inference. 

We experiment on the 12 multi-label datasets that are shown in Table \ref{tbl:datasets}. We adopt a 10-fold cross-validation process and discuss the results in terms of mean average precision (MAP) across all labels, as this was also the measure of choice in the ImageCLEF 2011 challenge that motivated this work. MAP is also the standard evaluation measure in multimedia information retrieval. 
%could remove strafication mentioning

\begin{table}[ht]
\setlength{\tabcolsep}{3pt}
\caption{A variety of multi-label datasets and their statistics: number of 
labels, examples, discrete and continuous features, and the mean number of inferring and excluding constraints that were discovered in the training set.}
\centering
%\scriptsize
\begin{tabular}{cccccccc}
\hline\noalign{\smallskip}
        &			& &			& \multicolumn{2}{c}{variables} &	\multicolumn{2}{c}{entailment}			  \\
dataset & source & labels 	& examples  & disc.  & cont. & positive & exclusion \\
\hline\noalign{\smallskip}
Bibtex & \cite{katakis08} 			& 159 	& 7395 		& 1836 	& 0		&11$\pm$0	& 76.2 $\pm$ 2.3\\
Bookmarks & \cite{katakis08} 		& 208 	& 87856 	& 2150	& 0   	&4.1 $\pm$ 0.3 & 1 $\pm$ 0\\
Emotions & \cite{trohidis+etal:2008}& 6 	& 593 		& 0 	& 72 	&0	&1.1$\pm$0.3\\
Enron	& url\tablefootnote{\url{http://bailando.sims.berkeley.edu/enron\_email.html}}		 					& 53 	& 1702 		& 1001 	& 0   	& 13 $\pm$ 15.2	& 480.7 $\pm$ 98.4\\
ImageCLEF2011 &	\cite{nowak+etal:2011}	& 99 	& 8000		& 19540 & 0   	&27.9 $\pm$ 0.9	& 325.4 $\pm$ 31.9  \\
ImageCLEF2012 &	\cite{thomee+popescu:2012}	& 94 	& 15000 	& 10000 & 0   	&1.2	& 277.9 $\pm$ 43.5 \\
IMDB & \cite{read:2011} 							& 28 	& 120919 	& 0 	& 1001  &0	& 21.6 $\pm$ 1.2\\
Medical & \cite{pestian+etal:2007} 	& 45 	& 978  		& 1449 	& 0   	&$6.3$ $\pm$ $1$ & $30.7$ $\pm$ $7.2$ \\
Scene & \cite{boutell-etal-2004}	& 6 	& 2407 		& 0 	& 294	&0	&4 $\pm$ 0\\
Slashdot  & \cite{read:2011}						& 20 	& 3782 		& 0 	& 1079	&0 &23.2 $\pm$ 1.2\\
TMC2007 & \cite{srivastava05} 		& 22 	& 28596 	& 49060	& 0 	&0	&7.5 $\pm$ 1.1\\
Yeast & \cite{elisseeff-weston-2002}& 14 	& 2417 		& 0    	& 103 	& 3 $\pm$ 0		&2.3 $\pm$ 0.5\\
\hline
\end{tabular}
\label{tbl:datasets}
\end{table}

We set the {\em minimum support} of discovered relationships to just 2 training examples (avoid single-point generalization). For positive entailment, {\em support} refers to the positive training examples of the antecedent label, while for exclusion it refers to the sum of the positive examples of all participating labels ($S$ and $T+U$ in Figure \ref{fig:contingency} respectively). In Section \ref{sec:future}, we discuss ideas for improved support setting. In 3 datasets (Bibtex, Bookmarks, Medical) exclusion discovery did not finish within a week, while in 3 other datasets (Enron, ImageCLEF2011/2012) a large number of exclusion rules was discovered that caused memory outage during network construction in jSMILE. We increased the support exponentially (4, 8, 16, ...) until these issues were resolved. 

\subsection{Relationships}

This section discusses the relationships discovered by our approach. At each fold of the cross-validation, different relationships can be discovered. Table \ref{tbl:datasets} reports the mean of the discovered relationships across all folds. We only discuss here those appearing in all folds. The relationships of {\em Yeast} and {\em TMC2007} cannot be discussed, as we were unable to determine their label semantics. Due to space limitation, we refrain from discussing less noteworthy relationships ({\em Slashdot}). %Tables presenting positive entailment relationships include a column mentioning the relationship support.   

\subsubsection{Bibtex and Bookmarks.}

The labels of the {\em Bibtex} and {\em Bookmarks} datasets correspond to tags assigned to publications and bookmarks respectively by users of the social bookmark and publication sharing system Bibsonomy\footnote{\url{http://www.bibsonomy.org/}}. 

Table \ref{tbl:bibtex-entailment} presents the 11 positive entailments that were found in all folds for {\em Bibtex}. These apparently correspond to a hierarchy relationship between label {\em statphys23}, an international conference on statistical physics\footnote{\url{http://www.statphys23.org/}} and the 11 topics of this conference. %It could be that the user(s) that added publications from this conference renamed label {\em topic5} to the more descriptive {\em nonequilibrium}, but did not bother for the rest of the topics. However, on the conference site, topic 5 is listed as {\em Dynamical systems and turbulence}, while {\em Nonequilibrium systems} is topic 3 of the conference. 
Table \ref{tbl:bookmarks-entailment} presents the 4 positive entailments that were discovered in all folds for {\em Bookmarks}. The first two most probably belong to a single user who also used the tags {\em film} and {\em kultur} whenever he/she used the tag {\em filmsiveseenrecently}. This is, by the way, an example of an unfortunate choice of tag name, as it involves a time adverb {\em recently}, whose meaning changes over time. The last two are examples of discovered is-a relationships, as paddling {\em is-a} (water) sport. We conclude that our approach manages to discover positive entailment relationships of {\em social} origin. 

\begin{table}
\setlength{\tabcolsep}{1.5pt}
\centering
\caption{Positive entailment relationships discovered in the {\em Bibtex} dataset.}
\begin{scriptsize}
\begin{tabular}{crclc|crclc|crclc}
\hline\noalign{\smallskip}
id & \multicolumn{3}{c}{relationship} & sup & id & \multicolumn{3}{c}{relationship} & sup & id & \multicolumn{3}{c}{relationship} & sup \\
\hline\noalign{\smallskip}
1 & nonequilibrium & $\rightarrow$ & statphys23 & 68 & 5 & topic4 & $\rightarrow$ & statphys23 & 62 & 9 & topic9 & $\rightarrow$ & statphys23 & 82 \\
2 & topic1 & $\rightarrow$ & statphys23 & 86 & 6 & topic6 & $\rightarrow$ & statphys23 & 63 & 10 & topic10 & $\rightarrow$ & statphys23 & 130 \\
3 & topic2 & $\rightarrow$ & statphys23 & 75 & 7 & topic7 & $\rightarrow$ & statphys23 & 129 & 11 & topic11 & $\rightarrow$ & statphys23 & 143 \\
4 & topic3 & $\rightarrow$ & statphys23 & 151 & 8 & topic8 & $\rightarrow$ & statphys23 & 73 \\
\noalign{\smallskip}\hline
\end{tabular}
\end{scriptsize}
\label{tbl:bibtex-entailment}
\end{table}

\begin{table}
\setlength{\tabcolsep}{4pt}
\centering
\caption{Positive entailment relationships discovered in the {\em Bookmarks} dataset.}
\begin{scriptsize}
\begin{tabular}{crclc|crclc}
\hline\noalign{\smallskip}
id & \multicolumn{3}{c}{relationship} & sup & id & \multicolumn{3}{c}{relationship} & sup \\
\hline\noalign{\smallskip}
1 & filmsiveseenrecently & $\rightarrow$ & film & 370 & 3 & paddling & $\rightarrow$ & sports & 379\\
2 & filmsiveseenrecently & $\rightarrow$ & kultur & 370 & 4 & paddling & $\rightarrow$ & watersports & 379\\
\noalign{\smallskip}\hline
\end{tabular}
\end{scriptsize}
\label{tbl:bookmarks-entailment}
\end{table}

A minimum support of 128 and 2048 examples led to a mean number of 76.2 and 1 exclusion relationships per fold in {\em Bibtex} and {\em Bookmarks} respectively. Due to space limitations we refrain from reporting the 18 exclusions discovered in all folds of {\em Bibtex}. In {\em Bookmarks} the relationship involved the following pair of labels: \{{\em computing}; {\em video}\}. No wonder it did not help improve accuracy. 

% discussion of excluding for bookmarks pending

\subsubsection{Emotions.} 

The 6 labels in the {\em Emotions} dataset concern 3 pairs of opposite emotions of the Tellegen-Watson-Clark model of mood: ({\em quiet-still}, {\em amazed-surprised}), ({\em sad-lonely}, {\em happy-pleased}) and ({\em relaxing-calm}, {\em angry-aggresive}) that correspond to the axes of engangement, pleasantness and negative affect respectively. The Tellegen-Watson-Clark model includes a fourth axis that concerns positive affect. The single discovered exclusion relationship, concerns the pair of opposite labels related to engangement: \{{\em quiet-still}; {\em amazed-surprised}\}. 
%\cite{tellegen+etal:1999}

\subsubsection{Enron.} 

The 53 labels of this dataset are organized into 4 categories: {\em coarse genre}, {\em included/forwarded information}, {\em primary topics}, which is applicable if coarse genre {\em Company Business, Strategy, etc} is selected, and {\em emotional tone} if not neutral. There are 13 positive entailment relationships by definition, as there are 13 labels in the {\em primary topics} category, which are children of label {\em Company Business, Strategy, etc}. 

Table \ref{tbl:enron-entailment} presents the 3 positive entailment relationships that were discovered in all folds. Relationship 1 is among the 13 positive entailments we already knew from the description of the labels, as label {\em company image -- changing / influencing} is a primary topic and therefore a child of label {\em Company Business, Strategy, etc}. Our approach manages to discover explicit is-a relationships, when these are present in the training data.

\begin{table}
\setlength{\tabcolsep}{1pt}
\centering
\caption{Positive entailment relationships discovered in the {\em Enron} dataset.}
\begin{scriptsize}
\begin{tabular}{crclc}
\hline\noalign{\smallskip}
id & \multicolumn{3}{c}{relationship} & sup \\
\hline\noalign{\smallskip}
1 & company image -- changing / influencing & $\rightarrow$ & Company Business, Strategy, etc. & 63 \\
2 & triumph / gloating & $\rightarrow$ & Company Business, Strategy, etc. & 3\\
3 & triumph / gloating & $\rightarrow$ & regulations and regulators (includes price caps) & 3\\
\noalign{\smallskip}\hline
\end{tabular}
\end{scriptsize}
\label{tbl:enron-entailment}
\end{table}

A minimum support of 8 examples led to a mean number of 480.7 exclusion relationships per fold. Only one relationship was present in all folds and involved the following interesting pair of concepts \{{\em Company Business, Strategy, etc}; {\em friendship / affection}\}. The conclusion is that there is no room for affection in the business world. 

\subsubsection{ImageCLEF 2011 and 2012.}

Labels in these two datasets correspond to 99 and 94 concepts respectively covering a variety of concepts for image annotation. A difference in the 2012 version of the contest was that concepts being superclasses of other concepts (e.g. {\em Animal}, {\em Vehicle} and {\em Water}) were removed. This is why in the 2011 version of the dataset 27 positive entailments were found in all folds (see Table \ref{tbl:ic2011-entailment}), in contrast with the following single one in the 2012 version: {\em Spider} $\rightarrow$ {\em QuantityNone}, with a support of 16 examples. The consequent of this relationship refers to the number of people that appear in the photo. %In other words, no people appear in the 16 spider pictures of that photo collection. 

\begin{table}
\setlength{\tabcolsep}{1.5pt}
\centering
\caption{Positive entailment relationships discovered in the {\em ImageCLEF2011} dataset.}
\begin{scriptsize}
\begin{tabular}{crclc|crclc|crclc}
\hline\noalign{\smallskip}
id & \multicolumn{3}{c}{relationship} & sup & id & \multicolumn{3}{c}{relationship} & sup & id & \multicolumn{3}{c}{relationship} & sup \\
\hline\noalign{\smallskip}
1 & Desert & $\rightarrow$ & Outdoor & 30 		& 10 & Sea & $\rightarrow$ & Outdoor & 222 	& 19 & Fish & $\rightarrow$ & Animals & 25 \\
2 & Desert & $\rightarrow$ & Day & 30 			& 11 & Sea & $\rightarrow$ & Water & 222 	& 20 & Bicycle 	  & $\rightarrow$ & NeutralLight & 61 \\
3 & Spring & $\rightarrow$ & NeutralLight & 105 & 12 & Cat & $\rightarrow$ & Animals & 61 	& 21 & Bicycle 	  & $\rightarrow$ & Vehicle & 61 \\
4 & Flowers & $\rightarrow$ & Plants & 367 		& 13 & Dog & $\rightarrow$ & Animals & 211  & 22 & Car 		  & $\rightarrow$ & Vehicle & 268 \\
5 & Trees & $\rightarrow$ & Plants & 890 		& 14 & Horse & $\rightarrow$ & Outdoor & 28 & 23 & Train   	  & $\rightarrow$ & Vehicle & 59 \\
6 & Clouds & $\rightarrow$ & Sky & 1104 		& 15 & Horse & $\rightarrow$ & Day & 28 	& 24 & Airplane	  & $\rightarrow$ & Vehicle & 41 \\
7 & Lake & $\rightarrow$ & Outdoor & 89 		& 16 & Horse & $\rightarrow$ & Animals & 28 & 25 & Skateboard & $\rightarrow$ & Vehicle & 12 \\
8 & Lake & $\rightarrow$ & Water & 89 			& 17 & Bird & $\rightarrow$ & Animals & 183	& 26 & Ship       & $\rightarrow$ & Vehicle & 79 \\
9 & River & $\rightarrow$ & Water & 130 		& 18 & Insect & $\rightarrow$ & Animals & 91& 27 & Female     & $\rightarrow$ & Male & 1254 \\
\noalign{\smallskip}\hline
\end{tabular}
\end{scriptsize}
\label{tbl:ic2011-entailment}
\end{table}

% discussion of excluding

A minimum support of 32 and 64 examples led to a mean number of 325.4 and 277.9 exclusion relationships per fold in the 2011 and 2012 version of {\em ImageCLEF} respectively. Due to space limitations we refrain from reporting the 24 and 21 exclusion relationships that were discovered in all folds of the 2011 and 2012 version of {\em ImageCLEF} respectively.

\subsubsection{IMDB.}

Labels of this dataset correspond to 28 movie genres. Table \ref{tbl:imdb-exclusion} presents the 15 exclusion relationships that were found in all folds. Labels {\em Film-Noir}, {\em Game-Show} and {\em Talk-Show} and are the most frequent ones in these relationships. We notice some obvious exclusions, such as \{{\em War}, {\em Reality-TV}\}, \{Game-Show, Crime\} and \{{\em Talk-Show}, {\em Fantasy}\}. On the other hand, such relationships could be a source of inspiration for innovative (or provocative) producers and directors contemplating unattempted combinations of genres.

\begin{table}
\setlength{\tabcolsep}{2pt}
\centering
\caption{Exclusion relationships discovered in the {\em IMDB} dataset.}
\begin{scriptsize}
\begin{tabular}{p{12cm}}
\hline\noalign{\smallskip}
\{Film-Noir, Game-Show, Adult, News\}, \{Film-Noir, Adult, Family\}, \{Game-Show, Horror\} \\
\{Film-Noir, Game-Show, Western\}, \{Film-Noir, Documentary\}, \{Game-Show, Thriller\} \\
\{Film-Noir, Talk-Show, Western\}, \{Talk-Show, Adventure\}, \{Game-Show, Crime\}  \\
\{Game-Show, Adult, Biography\}, \{Film-Noir, Comedy\}, \{War, Reality-TV\} \\
\{Film-Noir, Western, Reality-TV\}, \{Talk-Show, Mystery\}, \{Talk-Show, Action\} \\
\noalign{\smallskip}\hline
\end{tabular}
\end{scriptsize}
\label{tbl:imdb-exclusion}
\end{table}

\subsubsection{Medical.}

Labels of this dataset correspond to 45 codes/descriptions of the 9th revision of the International Statistical Classification of Diseases (ICD). Table \ref{tbl:medical-entailment} presents the 3 positive entailment relationships that were found in all folds. The support of these relationships is quite weak (up to 4 examples), yet it apparently corresponds to valid, yet already known,  medical knowledge. For example, the top page returned by Google for the query ``hydronephrosis congenital obstruction of ureteropelvic junction", contains the following excerpt: {\em Ureteropelvic junction obstruction is the most common pathologic cause of antenatally detected hydronephrosis}. Future work could apply our approach to larger datasets in search of unknown medical knowledge.   

\begin{table}
\setlength{\tabcolsep}{1pt}
\centering
\caption{Positive entailment relationships discovered in the {\em Medical} dataset.}
\begin{scriptsize}
\begin{tabular}{cp{6.65cm}cp{4.31cm}c}
\hline\noalign{\smallskip}
id & \multicolumn{3}{c}{relationship} & sup \\
\hline\noalign{\smallskip}
1 & 753.21 Congenital obstruction of ureteropelvic junction & $\rightarrow$ & 591 Hydronephrosis & 4 \\
2 & 786.05 Shortness of breath & $\rightarrow$ & 753.0 Renal agenesis and dysgenesis & 4\\
3 & 787.03 Vomiting alone & $\rightarrow$ & 753.0 Renal agenesis and dysgenesis & 3\\
\noalign{\smallskip}\hline
\end{tabular}
\end{scriptsize}
\label{tbl:medical-entailment}
\end{table}

A minimum support of 16 examples led to a mean number of 30.7 exclusion relationships per fold. Only one relationship was present in 9 out of the 10 folds (none in all folds) and involved the following 5 concepts: \{753.0 Renal agenesis and dysgenesis; 599.0 Urinary tract infection, site not specified; 596.54 Neurogenic bladder NOS; 780.6 Fever and other physiologic disturbances of temperature regulation; 493.90 Asthma,unspecified type, unspecified\}.

\subsubsection{Scene.}

Labels in this dataset correspond to 6 different scenery concepts. The following 2 exclusion relationships were found in all folds: \{Sunset, Fall Foliage, Beach\} and \{Sunset, Fall Foliage, Urban\}. These relationships do not really correspond to interesting knowledge, as images with such concept combinations can be found on the Web. However, only a few of the top 30 images returned by Google image search do cover all three concepts of these relationships, a fact that highlights the co-occurence rarity of these concepts. Still, this knowledge should not be generalized beyond the particular limited collection of 2407 images from the stock photo library of Corel. 

%\subsubsection{Slashdot.}

%Labels in this dataset correspond to 20 topical categories of news articles. The following 4 exclusion relationships were found in all folds: \{Interviews, Apache, News, BSD, Idle, AskSlashdot\}, \{Interviews, Apache, Search, BSD, Idle, AskSlashdot\}, \{Apache, Search, Science, BookReviews, Linux\} and \{Apache, Search, BSD, Games, BookRevies\}. This knowledge does not seem particularly interesting. 

%\subsubsection{TMC2007.} Labels in this dataset correspond to problems that might occur during flights and come from the database of NASA's aviation safety reporting system. Unfortunately, we could not retrieve the actual label semantics. 

%\subsubsection{Yeast.} Labels in this dataset correspond to 14 particular gene functional classes, but unfortunately we do not know the one-to-one match of these functional classes with the variables in the dataset. Personal communication on this issue with the authors of \cite{elisseeff-weston-2002} did not resolve the problem, despite their positive response and effort to help\footnote{If you happen to know the actual labels, please communicate them to us.}. 
%from the FunCat hierarchy \cite{ruepp+etal:2004},

\subsection{Results}

Tables \ref{tbl:MAPinferring} and \ref{tbl:MAPexcluding} present the average MAP of BR across the 10 folds of the cross-validation: (i) in its standard version, and (ii) with the exploitation of {\em positive entailment} relationships (Table \ref{tbl:MAPinferring}) and {\em exlusion} relationships (Table \ref{tbl:MAPexcluding}) via our approach. They also present the percentage of improvement brought by our approach, the minimum support and the average number of discovered relationships across the 10 folds of the cross-validation.  

\begin{table}
\setlength{\tabcolsep}{3pt}
\centering
\caption{Mean MAP of BR with and without exploitation of positive entailments via our approach, along with the percentage of improvement, the minimum support and the average number of discovered relationships.}
\begin{tabular}{cccccc}
\hline\noalign{\smallskip}
dataset  & standard BR & minsup & positive entailment & impr\% & \#relations \\
\hline\noalign{\smallskip}
Bibtex & 0.2152 $\pm$ 0.0114 & 2 & 0.2168 $\pm$ 0.0109 & 0.279 & 11 $\pm$ 0\\ 
Bookmarks & $0.1474$ $\pm$ $0.0041$ & 2 & $0.1475$ $\pm$ $0.0041$ & 0.068 & 4.1 $\pm$ 0.3\\ 
Enron & $0.2810$ $\pm$ $0.0476$ & 2 & $0.2821$ $\pm$ $0.0480$ & 0.391 & 3.8 $\pm$ 0.9 \\ 
ImageCLEF2011 & $0.2788$ $\pm$ $0.0113$ & 2 & $0.2871$ $\pm$ $0.0100$ & 2.977 & $27.9$ $\pm$ $0.9$ \\ 
ImageCLEF2012 & $0.2376$ $\pm$ $0.0089$ & 2 & $0.2380$ $\pm$ $0.0084$ & 0.168 & $1.2$ $\pm$ $0.9$ \\ 
Medical & $0.5997$ $\pm$ $0.0768$  & 2 & $0.6134$ $\pm$ $0.0661$ & 2.284 & 6.3 $\pm$ 1 \\
Yeast & $0.4545$ $\pm$ $0.0145$ & 2 & $0.4617$ $\pm$ $0.0141$ & 1.584 & 3 $\pm$ 0\\ 
\hline
\end{tabular}\\
\label{tbl:MAPinferring}
\end{table}

In Table \ref{tbl:MAPinferring} we notice that in all datasets where positive entailments were discovered, the exploitation of these relationships led to an increased MAP. Applying the Wilcoxon signed rank test we find a p-value of 0.0156, indicating that the the improvements are statistically significant. In some datasets, such as {\em bookmarks}, improvements are small, while in others, such as {\em ImageCLEF2011}, improvements are large. The correlation coefficient between the percentage of improvement and the number of relationships divided by the number of labels is 0.913, which supports the argument that the more relationships we discover per label, the higher the improvements in MAP. This was expected to an extend as MAP is a mean of the average precision across {\em all} labels. Focusing only on the affected labels, we would notice larger improvements. 

\begin{table}
\setlength{\tabcolsep}{3pt}
\centering
\caption{Mean MAP of BR with and without exploitation of exlusions via our approach, along with the percentage of improvement, the minimum support and the average number of discovered relationships.}
\begin{tabular}{cccccc}
\hline\noalign{\smallskip}
dataset  & standard BR & minsup & exclusion & impr\% & \#relations \\
\hline\noalign{\smallskip}
Bibtex & 0.2152 $\pm$ 0.0114 & 128 & 0.2117 $\pm$ 0.0126 & -1.626 & 76.2 $\pm$ 2.3\\ 
Bookmarks & $0.1474$ $\pm$ $0.0041$ & 2048 & $0.1473$ $\pm$ $0.0040$ & -0.068 & 1 $\pm$ 0\\ 
Emotions & $0.7163$ $\pm$ $0.0339$& 2 & $0.7265$ $\pm$ $0.0368$ & 1.424 & 1.1 $\pm$ 0.3 \\
Enron & $0.2810$ $\pm$ $0.0476$ & 8 & $0.2573$ $\pm$ $0.0476$ & -8.434 & 480.7 $\pm$ 98.4 \\ 
ImageCLEF2011 & $0.2788$ $\pm$ $0.0113$ & 32 & $0.2840 $ $\pm$ $0.0132$ & 1.865 & 325.4 $\pm$ 31.9 \\ 
ImageCLEF2012 & $0.2376$ $\pm$ $0.0089$ & 64 & $0.2308$ $\pm$ $0.0090$ & -2.862 & 277.9 $\pm$ 43.5 \\ 
IMDB & $0.0900$ $\pm$ $0.0030$& 2 & $0.0938$ $\pm$ $0.0026$ & 4.222 & 21.6 $\pm$ 1.2 \\ 
Medical & $0.5997$ $\pm$ $0.0768$ & 16 & $0.6223$ $\pm$ $0.0597$ & 3.769 & 30.7 $\pm$ 7.2 \\
Scene & $0.8139$ $\pm$ $0.0146$ & 2 & $0.8385$ $\pm$ $0.0140$ & 3.023 & 4 $\pm$ 0 \\ 
Slashdot & $0.3982$ $\pm$ $0.0323$& 2 & $0.4452$ $\pm$ $0.0422$ & 11.803 & 23.2 $\pm$ 1.2 \\ 
TMC2007 & $0.3276$ $\pm$ $0.0069$& 2 & $0.3474$ $\pm$ $0.0075$ & 6.044 & 7.5 $\pm$ 1.1\\ 
Yeast & $0.4545$ $\pm$ $0.0145$& 2 & $0.4625$ $\pm$ $0.0163$ & 1.760 & 2.3 $\pm$ 0.5\\ 
\hline\noalign{\smallskip}
Bibtex & 0.2152 $\pm$ 0.0114 & 256 & 0.2165 $\pm$ 0.0114 & 0.604 & 2.8 $\pm$ 0.4\\ 
Enron & $0.2810$ $\pm$ $0.0476$ & 32 & $0.2816$ $\pm$ $0.0477$ & 0.214 & 22.2 $\pm$ 1.8 \\ 
ImageCLEF2011 & $0.2788$ $\pm$ $0.0113$ & 128 & $0.2881 $ $\pm$ $0.0129$ & 3.336 & 56.6 $\pm$ 3.3 \\ 
ImageCLEF2012 & $0.2376$ $\pm$ $0.0089$ & 256 & $0.2391$ $\pm$ $0.0087$ & 0.631 & 39.8 $\pm$ 1.1 \\ 
\hline
\end{tabular}
\label{tbl:MAPexcluding}
\end{table}

\begin{figure}
\centering
\includegraphics[width=0.6\textwidth]{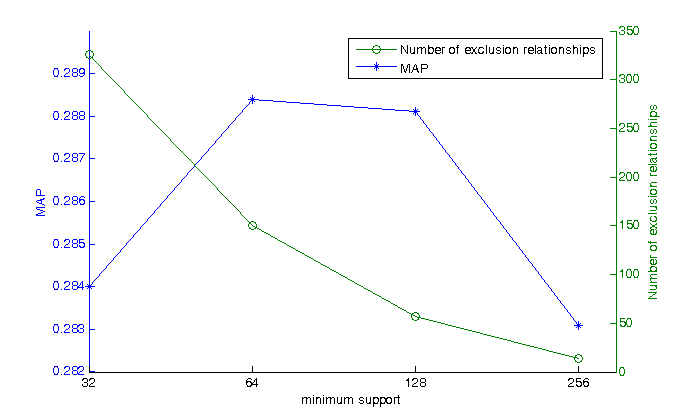}
\caption{\label{fig:ic2011}MAP and number of exclusion relationships for the 4 different minimum support values we tried in {\em ImageCLEF2011}.}
\end{figure}

In the upper part of Table \ref{tbl:MAPexcluding} we notice that there are 8 datasets where our approach leads to improvements in MAP, but another 4 where it leads to reductions. The p-value of the Wilcoxon signed rank test is 0.1099, indicating that the results are statistically insignificant even at the 0.1 level (though marginally). We argue that one of the main reasons underlying the negative results of our approach is the large number of spurious exclusion relationships that can be discovered in datasets with a larger number of labels. Indeed, the rarer two labels are, the higher the probability of being considered as mutually exclusive, irrespectively of their actual semantic relationship. We therefore further experiment with exponentially increasing minimum support values in {\em Bibtex}, {\em Enron} and the {\em ImageCLEF} datasets, where a large number of relationships was discovered, until we reach a small set of relationships. The bottom part of Table \ref{tbl:MAPexcluding} shows the results achieved through this process. We see that MAP improvements are now achieved for {\em Bibtex}, {\em Enron} and {\em ImageCLEF2012}, while the already improved MAP of {\em ImageCLEF2011} increases. Note that these are not the best achievable results, both because we only tried a few minimum support values and because our goal was to discover a small number of relationships. Figure \ref{fig:ic2011} shows the MAP and number of exclusion relationships for the minimum support values we tried in {\em ImageCLEF2011}. 

%add graph imagel clef 2011

The number of exclusion relationships divided by the number of labels is again positively correlated with the percentage of improvement, with the coefficient being 0.770 in this case (based on the bottom part of Table \ref{tbl:MAPexcluding} for the 4 datasets and ignoring {\em Bookmarks}). The coefficient would probably be higher had we calculated the actual number of labels involved in the exclusion relationships.

%it would be nice to also compute the average of the number of different labels involved in each set of exclusion constraints

The average improvement offered by exclusion (3.4\%) is larger than those offered by positive entailment (1.1\%), and so is the average number of relationships (19.3 vs 8.2). Exclusions typically involve more than two labels, while positive entailments are pairwise and some of them redundant due to the transitivity property of positive entailment. A deeper analysis should look at the number of labels involved in each type of relationship, which we leave as future work. 

Table \ref{tbl:both} shows results of utilizing both types of relationships. We notice that in {\em Yeast}, {\em Medical} and {\em ImageCLEF2012} the combination of both types of relationships leads to larger improvement than their individual improvement. The combined improvement is smaller than the sum of individual improvements. This could be due to labels appearing in both types of relationship. For {\em Enron} and {\em ImageCLEF2011} we notice that the combined improvement is smaller than the largest of the two individual improvements, that of positive entailment. This is another indication of spurious exclusions existence. Results on {\em Bibtex} were not obtained due to an error from jSMILE, which we are currently investigating. 

\begin{table}
\setlength{\tabcolsep}{3pt}
\centering
\caption{Mean MAP of BR with and without exploitation of both types of relationships via our approach, along with the percentage of improvement.}
\begin{tabular}{cccccc}
\hline\noalign{\smallskip}
         &             & \multicolumn{2}{c}{minimum support}   & &     \\     
dataset  & standard BR & positive & exlusion & our approach & impr\% \\
\hline\noalign{\smallskip}
Bookmarks & $0.1474$ $\pm$ $0.0041$   & 2 & 2048 & $0.1474$ $\pm$ $0.0040$ & 0 \\ 
Enron & $0.2810$ $\pm$ $0.0476$   & 2 & 8 & $0.2816$ $\pm$ $0.0477$ & 0.214 \\ 
ImageCLEF2011 & $0.2788$ $\pm$ $0.0113$   & 2 & 32 & $0.2846$ $\pm$ $0.0129$ & 2.080 \\ 
ImageCLEF2012 & $0.2376$ $\pm$ $0.0089$   & 2 & 256 & $0.2393$ $\pm$ $0.0085$ & 0.716 \\ 
Medical & $0.5997$ $\pm$ $0.0768$ & 2 & 16 & $0.6263$ $\pm$ $0.0566$ & 4.436 \\
Yeast & $0.4545$ $\pm$ $0.0145$   & 2 & 2 & $0.4677$ $\pm$ $0.0141$ & 2.904 \\ 
\hline
\end{tabular}\\
\label{tbl:both}
\end{table}

%\marginpar {Christina, can we find the actual rules and verify this?}

\section{Summary and Future Work}
\label{sec:future}

This work has introduced an approach that discovers entailment relationships among labels within multi-label datasets and exploits them using a sound probabilistic approach that enforces the adherence of the marginal probability estimates of multi-label learning with the discovered background knowledge.

We believe that our approach can be further extended and improved in a number of directions. An important issue concerns the statistical validity of the extracted relationships, especially when based on infrequent labels. We are working on automatically selecting the minimum support per relationship in order to separate chance artifacts from confident findings, which we expect not only to improve accuracy results, but also to reduce the complexity of the discovery process. On the opposite direction, it would be also interesting to investigate whether approximate relations, where the contingency table frequencies may not necessarily be zero due to noise, can be exploited with improved results. Another important direction is the generalization of our approach so as to be able to discover all types of entailments among any number of labels. 

On the empirical part of this work, we intend to apply our approach to additional datasets and to employ evaluation measures, such as log loss and squared error, that assess the quality of the predicted probabilities. We also intend to drill-down the accuracy results and discuss the extend of improvement for each label involved in one or more relationships. Finally, we also intend to investigate the effect that the quality of predicted probabilities has on our approach.

%\marginpar{stress the strength of binary relevance approaches}

\end{document}